\documentclass[lettersize,journal]{IEEEtran}
\usepackage{amsmath,amsfonts}
\usepackage{algorithmic}
\usepackage{algorithm}
\usepackage{array}
\usepackage[caption=false,font=normalsize,labelfont=sf,textfont=sf]{subfig}
\usepackage{textcomp}
\usepackage{stfloats}
\usepackage{url}
\usepackage{verbatim}
\usepackage{graphicx}
\usepackage{cite}
\usepackage{amsmath,amssymb}

\usepackage{xcolor}
\usepackage{float}

\usepackage{booktabs}
\usepackage{tabularx}
\usepackage{multirow}

\usepackage{tikz}
\usetikzlibrary{shapes.geometric, arrows, shadows, positioning, calc}

\begin{document}

\title{
Event-Centric World Modeling with Memory-Augmented
Retrieval for Embodied Decision-Making
}

\author{Zhaowen~Fan, 
        Rongchao~Zhang, 
        and~Yunxiang~Han%
\thanks{This work was supported by the National Natural Science Foundation of
China (No. 62273244).}
\thanks{Zhaowen Fan, Rongchao Zhang, and Yunxiang Han are with the School of
Computer Science, Sichuan University, Chengdu 610065, China (e-mail:
iamsuperfan@stu.scu.edu.cn; 2024141460297@stu.scu.edu.cn; hanscu@scu.edu.cn).}%
\thanks{Corresponding author: Yunxiang Han.}}

\maketitle

\begin{abstract}
Autonomous agents operating in dynamic environments increasingly demand
decision-making systems that are both efficient and interpretable. Hence we
propose Event-Retrieve-Action (ERA) framework, an alternative formulation for
embodied decision-making, bridging the gap between black-box imitation and
interpretable memory retrieval, while enabling online refinement without
retraining. The environment is represented as structured semantic events that are
encoded into an interpretable latent representation, and decisions are generated
by retrieving relevant prior experiences from a knowledge bank of event-action
pairs. Final actions are produced through weighted aggregation of retrieved
maneuvers, enabling transparent and physically consistent decision-making.
Experiments in UAV navigation are conducted to demonstrate real-time performance
and adaptive behaviors in dynamic environments as a representative embodied
decision-making application scenario.
\end{abstract}


\section{Introduction}
\label{introduction}

Autonomous systems, such as Unmanned Aerial Vehicles (UAVs) and Advanced Driver
Assistance Systems (ADAS), are increasingly deployed in complex and dynamic
environments where reliable and timely decision-making is critical and
\cite{thrun2002probabilistic}. In such settings, agents are expected not only to
react to sensory inputs, but also to reason about the evolving state of the
environment \cite{russell1995modern}. This has motivated the development of
predictive world models that enable agents to anticipate future states and plan
accordingly \cite{ha2018world, li2025comprehensive}.

Recent advances in deep learning have enabled end-to-end mappings from sensory
inputs to control actions, while recent world models introduce generative
predictive capabilities. Vision-Language-Action (VLA) architectures
\cite{zitkovich2023rt, o2024open} improve generalization across robotic tasks,
and newer variations incorporate implicit predictive mechanisms to reason about
physical dynamics \cite{zheng2025flare}, while diffusion-based methods
\cite{chi2025diffusion, yang2023unisim} excel at capturing multi-modal
distributions in imitation learning. However, these methods often impose
significant computational overhead, especially when combined with some complex
perception and downstream action layers, challenging the sub-10ms control loops
required for time-critical scenarios. Furthermore, end-to-end models often rely
on implicit representations embedded in neural parameters that are opaque and lack
interpretability and therefore are unable to provide explicit safety guarantees
required in safety-critical systems and might lead to unexplainable spurious
actions under distribution shift. As a result, it remains difficult to analyze
decision-making processes for debugging and optimization or verify whether
generated actions are consistent with underlying physical constraints
\cite{raissi2019physics, khalil2002nonlinear}, which is particularly problematic
in safety-critical applications.

To bridge the gap between expressive reasoning and physical consistency,
subsequent works have explored Physics-Informed Neural Networks (PINNs)
\cite{raissi2019physics}, structured representations \cite{zaheer2017deep},
model-based reasoning \cite{ha2018world}, and memory-augmented learning
\cite{graves2016hybrid}. While those works have emerged as a promising
alternative, both the representations and the underlying reasoning process
are still implicit, and the actions are still not possible to seamlessly debug
and optimize due to limited transparency and explicit verification. Moreover, the
physics in physics-guided and structure neural methods mainly influences learning
or latent representation, ignoring the core goal: consistency in embodied actions.
Consider a robotic manipulator retrieving a previously unseen object, a
physics-informed controller may satisfy the physics-guided governing equations
encoded during training, yet still generate a grasp trajectory that violates
collision constraints or object stability, while offering no explicit mechanism
to predict or verify these failures prior to execution nor identify which internal
representation or reasoning step led to the unsafe action.

Abover all, this raises a fundamental question: Must embodied world models
compress knowledge into neural parameters? We argue that such compression is not
the only viable paradigm; their limitations suggest that an alternative paradigm
deserves investigation. Intriguingly, the success of Retrieval-Augmented
Generation (RAG) in natural language processing has demonstrated that
externalizing knowledge into a queryable memory bank improves both factual
accuracy and model transparency \cite{lewis2020retrieval}. Additionally,
case-based reasoning (CBR) \cite{aamodt1994case} provides another perspective in
which decisions are derived from previously observed experiences, offering an
inherently interpretable mechanism for action selection. This observation
motivates an alternative perspective: explicit retrieval and reasoning may provide
an equally viable foundation for embodied decision-making without compressing all
knowledge into neural parameters. However, integrating such approaches into modern
embodied systems remains challenging, especially in dynamic environments with
multiple interaction entities, since such systems must ensure that retrieved
experiences are kinematically feasible and asymptotically stable under closed-loop
control. We are optimistic that it is possible to overcome these fundamental and
practical problems.

In this work, we introduce an Event-Retrieve-Action (ERA) framework for embodied
decision-making. Motivated by previous observations above, ERA adopts an
alternative philosophy in which prior experiences remain explicitly represented
rather than compressed into neural parameters. The key idea is to represent the
environment as a structured set of semantic events, capturing both object-level
properties and their dynamics together with the agent state. These event
representations are encoded into a latent space and used to query a knowledge
bank of prior experiences \cite{zhu2024retrieval, lewis2020retrieval}. Unlike
end-to-end policies that compress demonstrations into fixed network parameters,
our proposed mechanism explicitly preserves prior experiences as reusable decision
primitives, and the decision-making is performed by retrieving and combining
relevant solutions, enabling the agent to leverage previously observed strategies
in a transparent and structured manner, positioning the framework as a system
where the Event-to-Action link is explicitly traceable through the retrieval
process. The proposed framework integrates three key conceptual ideas: (i)
event-centric semantic abstraction for representing dynamic environments, (ii)
memory-augmented retrieval for leveraging prior experiences, and (iii)
interpretable decision-making through explicit aggregation of retrieved solutions.
This methodology provides a unified perspective on representation, reasoning, and
control in embodied systems.

Existing retrieval-based embodied decision-making frameworks remain limited in
safety-critical navigation because they typically retrieve actions based solely on
state similarity without considering physical feasibility, multimodal ambiguity,
or closed-loop stability. As a result, retrieved actions may become dynamically
infeasible, average toward unsafe intermediate maneuvers, or degrade under
repeated deployment. To the best of our knowledge, ERA is among the first
retrieval-based navigation frameworks to jointly integrate event-centric
abstraction, multimodal action disambiguation, and stability-aware retrieval
filtering in a unified pipeline. Our experiment results suggest that the ERA
framework enables fast and interpretable retrieval-based decision-making,
achieving higher trajectory efficiency than optimization-based control and
stronger reliability than reinforcement learning baselines while maintaining
real-time performances on edge hardware. The remainder of this paper is organized
as follows: Section~II details the mathematical formulation of the ERA framework;
Section~III presents the experimental results, deployment efficiency, and
comparative analysis; Section~IV discusses the broader implications and
limitations of the ERA framework; and Section~V concludes the work.

\section{The ERA Framework}
\label{the-era-framework}

\subsection{Overview}
\label{overview}

\tikzstyle{sensor} = [
  rectangle, rounded corners, minimum width=3cm, minimum height=1cm,
  text centered, draw=black, fill=gray!10, drop shadow
]
\tikzstyle{process} = [
  rectangle, rounded corners, minimum width=3cm, minimum height=1cm,
  text centered, draw=black, fill=blue!5, drop shadow
]
\tikzstyle{latent} = [
  rectangle, rounded corners, minimum width=3cm, minimum height=1cm,
  text centered, draw=black, fill=green!10, drop shadow
]
\tikzstyle{bank} = [
  trapezium, trapezium left angle=75, trapezium right angle=105,
  minimum width=3cm, minimum height=0.8cm, text centered, draw=black,
  fill=orange!10, drop shadow
]
\tikzstyle{action} = [
  rectangle, rounded corners, 
  minimum width=3.5cm, minimum height=1.2cm, 
  text centered, draw=black, fill=red!10, drop shadow
]
\tikzstyle{arrow} = [thick,->,>=stealth]

\begin{figure}
\centering
\begin{tikzpicture}[node distance=1.5cm, scale=0.5, auto]
\node (raw) [sensor] {Raw Sensory Input};
\node (event) [process, below of=raw] {Event List $E_t$};
\node (enc) [latent, below of=event] {Latent Code $z_t = f(E_t)$};
\node (retrieval) [bank, below of=enc] {Knowledge Bank $M$};
\node (n1) [
  rectangle, draw, dashed, below of=retrieval,
  xshift=-1.2cm, minimum width=1cm
] {$w_1 a_1$};
\node (n2) [
  rectangle, draw, dashed,below of=retrieval,
  xshift=0cm, minimum width=1cm
] {$w_2 a_2$};
\node (nk) [
  rectangle, draw, dashed, below of=retrieval,
  xshift=1.2cm, minimum width=1cm
] {$w_k a_k$};
\node (action) [
  action, below of=n2, fill=red!10
] {Action $a_t = \sum w_i a_i$};

\draw [arrow] (raw) -- node[anchor=west] {Feature Ext.} (event);
\draw [arrow] (event) -- node[anchor=west] {Encoding} (enc);
\draw [arrow] (enc) -- node[anchor=west] {Query} (retrieval);
\draw [arrow] (retrieval.south) -- (n1.north);
\draw [arrow] (retrieval.south) -- (n2.north);
\draw [arrow] (retrieval.south) -- (nk.north);
\draw [arrow] (n1.south) -- (n1.south |- action.north);
\draw [arrow] (n2.south) -- (action.north);
\draw [arrow] (nk.south) -- (nk.south |- action.north);
\draw [arrow] (action.east) -- ++(1.5,0) |- node[
  anchor=south, pos=0.25, rotate=90
] {Status Codes} (retrieval.east);
\end{tikzpicture}
\caption{\textbf{The workflow of the proposed framework}, the proposed framework
models the environment at time $t$ as an event representation $E_t$, which is
mapped into a latent space via an encoding function $z_t = f(E_t)$.
Decision-making is performed by querying the knowledge bank with $z_t$ and
retrieving a small set of candidate maneuvers. These candidates are then scored,
filtered, and selected to produce the final action.}
\label{fig:Fig.1}
\end{figure}

Figure~\ref{fig:Fig.1} illustrates the overall workflow of the ERA framework.
The ERA framework consists of four integrated stages: (i) feature extraction,
(ii) event abstraction into a structured list $E_t$, (iii) latent encoding via
$f(\cdot)$, and (iv) retrieval-based decision-making using a structured
knowledge bank $M$. The system further incorporates a feedback mechanism where
execution outcomes are recorded as status codes to refine future decisions
through online refinement. This retrieval-and-selection process enables
interpretable and structured decision-making based on prior experiences.

\subsection{Data Representation}
\label{data-representation}

\subsubsection{Event List}
\label{event-list}

The event list consists of a set of event elements defined as:

\begin{equation}
e_t^i =
\big[
c_i,\;
\Delta p_i,\;
\Delta v_i,\;
v_{\mathrm{ego}},\;
\Delta g
\big]
\in \mathbb{R}^{13},
\end{equation}

Each event element $e_t^i$ encodes one detected entity relative to the ego agent:

\begin{itemize}
\item \textbf{entity-type code} $c_i$: a categorical indicator distinguishing
object classes (e.g., drone, bird, other).
\item \textbf{relative position} $\Delta p_i$: the position of the detected
entity with respect to the ego agent.
\item \textbf{relative velocity} $\Delta v_i$: the velocity of the detected
entity relative to the ego agent.
\item \textbf{ego velocity} $v_{\mathrm{ego}}$: the current velocity of the ego
agent.
\item \textbf{goal-relative direction or offset} $\Delta g$: a goal-conditioning
term that preserves navigational intent.
\end{itemize}

Rather than appending a separate global state vector, the ego-motion and
goal-conditioning terms are embedded directly into each event element. The latent
code $z_t$ is then obtained by permutation-invariant aggregation over the event
set.

\subsubsection{Event Code}
\label{event-code}

The latent event code $z_t$ defined above is a compact, permutation-invariant
embedding of the event list. Implemented with a DeepSets-style encoder
\cite{zaheer2017deep}, it captures the spatial, dynamic, and task-conditioned
relationships among detected entities and serves as the query representation for
retrieval from the knowledge bank.

\subsubsection{Event Dynamics}
\label{event-dynamics}

The latent event representation evolves according to a locally linear transition
model:

\begin{equation}
z_{t+1} = \Psi z_t + \Gamma a_t + \epsilon_t, \quad \rho(\Psi) < 1
\end{equation}

where $\Psi$ is the latent transition operator, $\Gamma$ maps an executed
maneuver back into the latent space, and $\epsilon_t$ captures residual
modeling uncertainty. This formulation provides a compact local approximation of
the environment dynamics while preserving a direct connection between retrieved
actions and their latent consequences.

To maintain stable long-horizon behavior, we enforce contractive latent
dynamics by constraining the transition operator such that $\rho(\Psi) < 1$
\cite{ogata2010modern}. In the current implementation, this is realized by
clamping the singular values of $\Psi$ after each update, preventing the latent
transition from becoming expansive. Candidate maneuvers are then screened with
the identity Lyapunov function $V(z)=\|z\|^2,$ and a transition is accepted
only if it satisfies the discrete-time stability criterion

\begin{equation}
\Delta V_t = \|z_{t+1}\|^2 - \|z_t\|^2 < 0.
\end{equation}

This stability filter rejects maneuvers that increase latent energy and helps
preserve recoverable motion in dense or high-risk encounters.

In addition, the latent space is trained to be kinematically informed, so that
distances $d_{\mathrm{phys}}(z_t,z_i)$ correlate with the suitability of
reusing stored maneuvers at the current state. The resulting environment
transition is modeled as

\begin{equation}
E_{t+1} \sim p(E_{t+1}\mid E_t,a_t),
\end{equation}

where observed transitions are logged for online refinement. During deployment,
retrieval generates candidate actions based on latent similarity, while the
transition model $(\Psi, \Gamma)$ evaluates their predicted consequences before
execution. In the current implementation, $\Gamma$ is kept fixed during online
refinement, while $\Psi$ is updated under stability constraints.

\subsubsection{Status Code}
\label{status-code}

For online refinement, the system logs transition tuples of the form

\begin{equation}
S_t = (E_t, a_t, r_t, E_{t+1})
\end{equation}

where $r_t$ denotes the observed reward or feasibility signal. These logged
transitions are used to evaluate executed decisions and to support online
adaptation of the latent dynamics model.

This transition log is conceptually distinct from the retrieval memory. The
knowledge bank stores compact retrieval entries $(z_i, a_i, \rho_i)$ for action
reuse, while transition tuples retain the observed before/after event information
needed for refinement.

\subsection{System Architecture}
\label{system-architecture}

\subsubsection{Event Triggering Mechanism}
\label{event-triggering-mechanism}

The tracker is activated based on an event trigger condition. In this work, the
trigger is defined as the detection of the relevant object or intruder within a
predefined spatial threshold using a perception module (e.g., computer
vision-based object detection or sensor-based proximity detection).
Specifically, the tracker is initiated when
$d_{\text{object}} < d_{\text{threshold}}$, where $d_{\text{object}}$ denotes the
distance between the ego agent and the detected entity. This condition ensures
that only safety-critical or decision-relevant events activate the tracking and
memory recording process.

Alternatively, semantic triggers (e.g., object classification indicating
potential collision risk) can also be incorporated to enhance robustness.

\subsubsection{Knowledge Bank}
\label{knowledge-bank}

The knowledge bank is a structured memory module that stores retrieval entries
of the form

\begin{equation}
M = \{(z_i, a_i, \rho_i)\}
\end{equation}

where $z_i$ is a latent event code, $a_i$ is the associated maneuver, and
$\rho_i$ is a retrieval reliability score. In the current implementation, newly
inserted entries are initialized with a uniform reliability value 1.0 and are 
subsequently down-weighted when later observations indicate unsafe or near-unsafe
outcomes (e.g., warning or collision events).

\subsubsection{Retrieval Mechanism}
\label{retrieval-mechanism}

Given the latent event code $z_t$, the system retrieves a top-$k$ candidate set
$\mathcal{N}_k(z_t)$ according to latent-space similarity:

\begin{equation}
\mathcal{N}_k(z_t) =
\{(z_i,a_i,\rho_i) \mid z_i \in \text{Top-}k(\operatorname{sim}(z_t,z_i))\}.
\end{equation}

Candidate weights are computed from both latent proximity and reliability,

\begin{equation}
w_i = \operatorname{softmax}\!\left(-d(z_t,z_i)/\tau + \log(\rho_i)\right),
\end{equation}

after which unstable candidates are rejected and the remaining actions are passed
to the clustered selection module.

\subsubsection{Multi-modal Action Selection}
\label{multi-modal-action-selection}

To prevent the "average-to-collision" failure mode, a known challenge in
multimodal imitation learning \cite{zhang2018deep}—where averaging two safe but
opposing maneuvers (e.g., turn left vs. turn right) results in an unsafe
intermediate action (e.g., go straight)—we implement a Clustered Weighted
Selection (CWS) strategy. Retrieved maneuvers are grouped by directional cosine
similarity. We calculate the aggregate weight for each cluster $c$ as
$W_c = \sum_{i \in \text{Cluster } c} w_i$. The final action is obtained by
selecting the highest-weight group and performing weighted averaging only within
that group.

\subsection{Algorithms}
\label{algorithms}

\subsubsection{Optimization and Inference}
\label{optimization-and-inference}

The framework operates in two stages. Offline, a virtual-potential-field teacher
generates demonstration pairs $(E_t, a_t^*)$ for encoder pretraining and
knowledge bank initialization. Online, the agent encodes the current event list,
retrieves a candidate set of prior maneuvers, executes a selected action, and
periodically updates the transition model and retrieval reliability from observed
transitions.

\begin{equation}
\mathcal{L}(\theta_1) = \lambda_m \mathcal{L}_{metric} + \lambda_i
\mathcal{L}_{\text{imitation}}
\end{equation}

Here, $\mathcal{L}_{metric} = \left| \|z_i - z_j\|_2 -
\mathcal{D}_{phys}(E_i, E_j) \right|$ is the physical interpretability
constraint that aligns latent similarity with physical state similarity, while
$\mathcal{L}_{\text{imitation}} = \|a_t - a_t^*\|^2$ matches the latent code to
the teacher action \cite{ross2011reduction}.

\begin{equation}
\mathcal{L}(\theta_2) = \lambda_p \mathcal{R}_{\text{phys}}(w, z) - \lambda_r
\mathcal{J}_{\text{perf}}
\end{equation}

Here, $\mathcal{R}_{\text{phys}}$ encourages latent-transition consistency during
online refinement by favoring retrievals that remain compatible with the learned
local dynamics, while $\mathcal{J}_{\text{perf}}$ optimizes task performance.
In the current implementation, $\mathcal{R}_{\text{phys}}$ primarily serves to
maintain structural consistency of the latent representation rather than directly
optimizing benchmark performance, and the encoder remains frozen during this
stage, while online refinement is applied to the transition model and retrieval
reliability.

\begin{algorithm}
\footnotesize
\caption{ERA Optimization and Inference}
\label{alg:era}
\begin{algorithmic}[1]
\REQUIRE Demonstration set $\mathcal{D}=\{(E_t,a_t^*)\}$, knowledge bank $M$
\ENSURE Action $a_t$

\STATE Pretrain encoder $f(\cdot)$ on $\mathcal{D}$ using $\mathcal{L}_{metric}$ and $\mathcal{L}_{\text{imitation}}$
\STATE Encode demonstrations and initialize the knowledge bank $M$

\FOR{each online step}
\STATE Detect event list $E_t$
\STATE Encode $z_t = f(E_t)$
\STATE Retrieve top-$k$ candidate maneuvers from $M$
\STATE Reject candidates that violate the latent stability constraint
\STATE Select final action $a_t$ by clustered weighted aggregation
\STATE Execute $a_t$ and observe $(r_t, E_{t+1})$
\STATE Store $(E_t, a_t, r_t, E_{t+1})$ and periodically update the transition model
\ENDFOR

\RETURN $a_t$
\end{algorithmic}
\end{algorithm}

\subsubsection{Time and Memory Complexity}
\label{time-and-memory-complexity}

The framework's computational overhead is dominated by retrieval, which scales
linearly with the size of the knowledge bank $|M|$ in the current
implementation, since the retrieve function computes distances from the query
latent code to all stored latent entries before applying top-$k$ selection on
the GPU. Encoding scales linearly with the number of detected objects $n$.
Action selection over the retrieved set is quadratic in $k$ in the current
implementation because pairwise cosine similarities are computed among the
retrieved candidates, but $k$ is small and fixed in practice.

Memory usage is linear in the number of stored experiences because the bank
keeps latent codes, actions, and reliability scores for each entry, giving
$\mathcal{O}(|M|\cdot d)$ storage. During training, only a temporary experience
buffer is added, so the main long-term cost remains the knowledge bank.

\section{Experiments and Results}
\label{experiments-and-results}

\subsection{Setup and Training Dynamics}
\label{setup-and-training-dynamics}

We evaluate ERA in NVIDIA Isaac Sim \cite{NVIDIA_Isaac_Sim} under a two-stage
training protocol consisting of offline expert initialization followed by online
adversarial refinement. The implementation is designed to preserve geometric
clarity and fast collision checking by modeling both the ego UAV and surrounding
agents as spherical primitives.

\subsubsection{Simulation Setup}
\label{simulation-setup}

All simulations are executed in Isaac Sim with a fixed physics step of
$\Delta t = 0.05\,\mathrm{s}$. The ego vehicle is initialized at
$\mathbf{p}_0 = [0, 0, 1.5]^\top$ and navigates toward a fixed goal at
$\mathbf{p}_g = [50, 20, 5]^\top$. The ego body is represented as a dynamic
sphere of radius $0.25\,\mathrm{m}$. Intruders are also instantiated as dynamic
spherical bodies, including drone-like moving obstacles, bird-like oscillatory
obstacles, and static obstacles. In the implementation, drone and static
obstacles use a radius of $0.30\,\mathrm{m}$, while bird-like intruders use a
smaller radius of $0.15\,\mathrm{m}$ to reflect their lighter kinematic profile.

At each control step, the environment is converted into an event list
$E_t = \{e_t^i\}_{i=1}^{N_t}$ containing only nearby intruders. Each event is
encoded as a 13-dimensional vector
$$
e_t^i =
[c_i,\ \Delta \mathbf{p}_i,\ \Delta \mathbf{v}_i,\ \mathbf{v}_{ego},\ \Delta \mathbf{g}],
$$
where $c_i \in \{0,1,2\}$ denotes the intruder type, $\Delta \mathbf{p}_i$ and
$\Delta \mathbf{v}_i$ are relative position and velocity with respect to the ego
vehicle, $\mathbf{v}_{ego}$ is the ego velocity, and
$\Delta \mathbf{g} = \mathbf{p}_g - \mathbf{p}_{ego}$ is the relative goal
vector. This event-centric representation provides the encoder with both local
interaction geometry and global navigation context.

The knowledge bank is bootstrapped from an expert dataset generated by a Virtual
Potential Field (VPF) teacher \cite{khatib1986real}. The teacher combines an
attractive force toward the goal with repulsive forces from nearby intruders,
and applies velocity clipping to enforce a maximum speed of
$5\,\mathrm{m/s}$. During dataset generation, up to five intruders are
maintained ahead of the ego vehicle, with spawn distances sampled from
$7$--$12\,\mathrm{m}$ and intercept-style velocities chosen to create
interaction-rich trajectories. We run 50 seeded rollout episodes of up to 1000
steps each and retain only successful goal-reaching trajectories. This produces
an expert dataset of 27,075 state-action pairs, which is then used to pretrain
the event encoder and initialize the knowledge bank.

Encoder pretraining uses a latent dimension of 128 and combines an imitation
loss with a metric-alignment loss so that distances in latent space remain
consistent with distances in the physical event space. After pretraining, each
expert pair $(E_t, a_t)$ is inserted into the knowledge bank together with a
synthetic next latent state obtained from a one-step kinematic update, thereby
bootstrapping retrieval before online refinement begins.

\subsubsection{Adversarial Curriculum}
\label{subsec:intruder_logic}

In the online refinement stage, ERA is exposed to denser and progressively
harder encounters through a curriculum-driven adversarial spawner
\cite{bengio2009curriculum, pinto2017robust}. In this phase, we are using a
different operating setting to avoid same distribution, such as sensing
threshold and intruder density, because the former is designed to generate clean
expert demonstrations, whereas the online refinement training is designed to
stress-test and refine the retrieval policy under denser conflict conditions.
In the simulator we maintain up to 25 active intruders and progressively
increases difficulty by reducing time-to-collision and shrinking
minimum-clearance margins. The difficulty variable is scheduled using a sigmoid
progression over training steps and is further adjusted by recent agent
performance, yielding an adaptive curriculum rather than a purely time-based
one.

The adversarial spawner synthesizes several encounter modes, including direct
collision-course approaches, near-miss events, crossing trajectories, and
multi-intruder conflict cases. Difficulty controls the encounter geometry by
interpolating the nominal time-to-collision from $2.5\,\mathrm{s}$ to
$0.25\,\mathrm{s}$ and the intended minimum clearance from $3.0\,\mathrm{m}$ to
$0.3\,\mathrm{m}$. Intruder approach speeds are clipped to the range
$5$--$8\,\mathrm{m/s}$, which produces aggressive but still kinematically
plausible interactions. This curriculum expands the retrieval distribution well
beyond the offline expert set and forces the knowledge bank to cover both common
and safety-critical events.

\subsubsection{Training Dynamics}
\label{subsec:training_dynamics}

\begin{figure}
    \centering
    \includegraphics[width=\columnwidth]{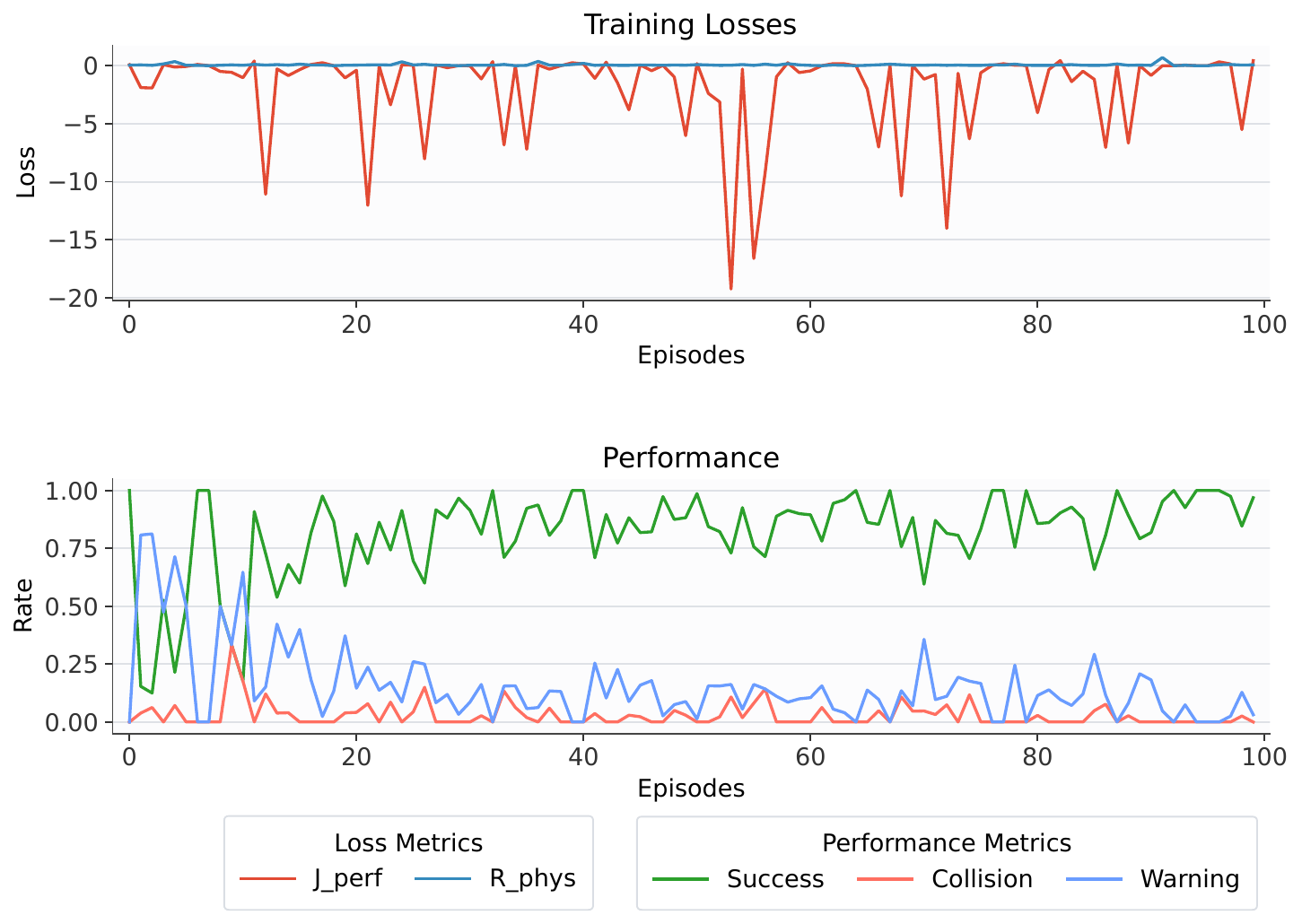}
    \caption{\textbf{Training dynamics during the first 100 episodes.} Left:
    online refinement losses, including the physics-consistency term $R_{phys}$
    and the retrieval-performance term $J_{perf}$. Right: episode-level success,
    collision, and warning rates under the adversarial curriculum.}
    \label{fig:Fig.2}
\end{figure}

During online refinement, action selection remains retrieval-based. For a query
latent state $z_t$, ERA retrieves the top-$k$ nearest memory entries
($k=5$), weights them by distance and memory reliability, filters them through a
stability check based on the latent transition model, and then aggregates only
the most self-consistent action cluster. The executed command is added to a
nominal goal-directed base velocity, so the policy preserves task progress even
when threat density is high.

The online update objective combines a physics-consistency term and a
performance-driven retrieval term. The physics term penalizes mismatch between
the predicted next latent state and the actually observed next latent state,
while the performance term encourages retrieval patterns associated with higher
reward. Rewards are shaped by goal progress and safety: goal completion receives
a large positive reward, cleared interactions receive a smaller positive reward,
unsafe proximity produces a negative penalty, and collisions receive a strong
negative penalty. In addition, memory entries associated with unsafe or
colliding actions have their reliability reduced, which makes the knowledge bank
self-correcting over time.

Fig.~\ref{fig:Fig.2} shows that, despite the increasing adversarial difficulty,
the physics-consistency loss remains bounded and the policy becomes progressively
safer. Success rate improves after early curriculum exposure, while collision and
warning events decrease overall, indicating that ERA can refine retrieval
behavior online without retraining the full model from scratch.

\subsection{System Validation}
\label{system-validation}

\subsubsection{Mathematical Fidelity and System Stability}
\label{mathematical-fidelity-and-system-stability}

\begin{figure}
    \centering
    \includegraphics[width=\columnwidth]{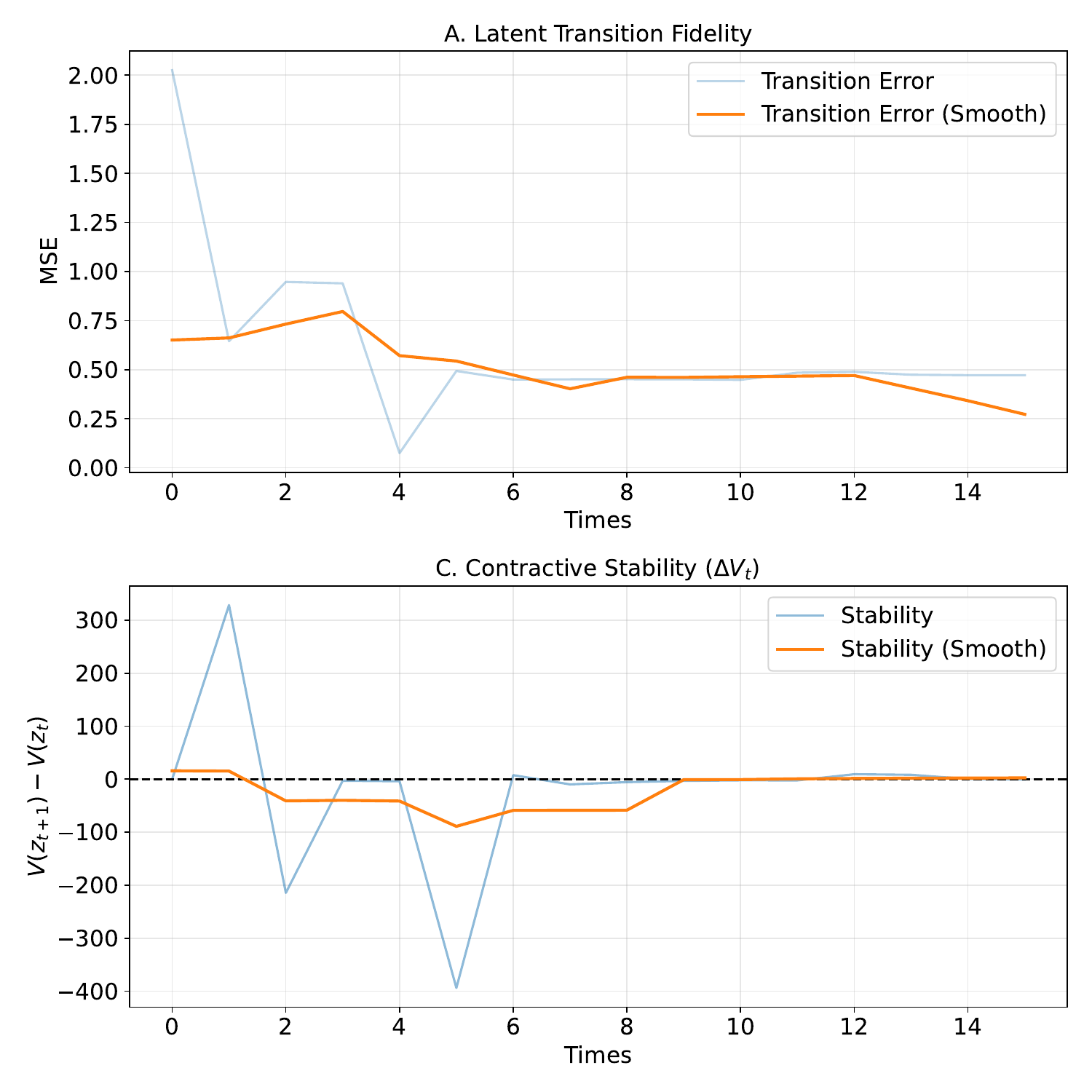}
    \caption{\textbf{Mathematical and Physical Validation.}
    \textit{(A) Latent Transition Fidelity:} One-step prediction error is
    evaluated by comparing the predicted latent successor
    $\hat{z}_{t+1} = z_t \mathbf{\Psi} + a_t \mathbf{\Gamma}^{\top}$ against the
    encoded successor state $z_{t+1}$. The mean per-step MSE is $0.61$, which
    corresponds to an explained variance of 84.8\% relative to the global latent
    variance ($\sigma^2 = 4.0275$), supporting the validity of the local linear
    transition model used during retrieval. \textit{(B) Latent Contraction:}
    Stability is evaluated through the Lyapunov energy
    $V(z)=\|z\|_2^2$ and its increment
    $\Delta V_t = V(z_{t+1}) - V(z_t)$. The dominance of negative
    $\Delta V_t$ excursions indicates that the closed-loop policy remains
    dissipative during conflict resolution rather than amplifying high-risk
    perturbations. Metrics are computed from a held-out seeded audit trajectory
    using a frozen checkpoint and frozen knowledge bank snapshot.}
    \label{fig:Fig.3}
\end{figure}

We audit the ERA framework's integrity via latent transition fidelity and
contraction analysis (Fig.~\ref{fig:Fig.3}). Unlike the training curves in
Fig.~\ref{fig:Fig.2}, this audit is performed with a frozen fine-tuned
checkpoint and a frozen knowledge bank snapshot on a held-out seeded deployment.
The goal is not to measure task success directly, but to verify that the
deployed retrieval controller remains internally consistent with the latent
transition model used by ERA. Concretely, we log the latent state $z_t$, the
retrieved action $a_t$, and the encoded successor state $z_{t+1}$ during
deployment, and then evaluate the one-step prediction
$\hat{z}_{t+1} = z_t \mathbf{\Psi} + a_t \mathbf{\Gamma}^{\top}$ against the
observed $z_{t+1}$ by per-step MSE. In parallel, we assess stability with the
Lyapunov energy $V(z)=\|z\|_2^2$ and the discrete energy increment
$\Delta V_t = V(z_{t+1}) - V(z_t)$.

This validation is important because ERA does not execute an end-to-end neural
policy; it executes a retrieved maneuver after latent filtering and stability
screening. Therefore, low transition error is evidence that the local linear
latent dynamics remain meaningful under closed-loop deployment, rather than only
on offline training samples. Likewise, predominantly negative $\Delta V_t$
values indicate that the selected maneuvers tend to contract the latent energy
instead of amplifying perturbations. As shown in Fig.~\ref{fig:Fig.3}, the
transition error remains small relative to the global latent variance, while the
energy trace repeatedly drops during conflict resolution. Taken together, these
results support the interpretation that the latent space is not merely
descriptive, but is sufficiently structured to support stable retrieval-guided
control under adversarial interactions.

\subsubsection{Knowledge Bank Scaling, Action Alignment, and Performance Saturation}
\label{knowledge-bank-scaling-and-performance-saturation}

\begin{figure*}
    \centering
    \includegraphics[width=\textwidth]{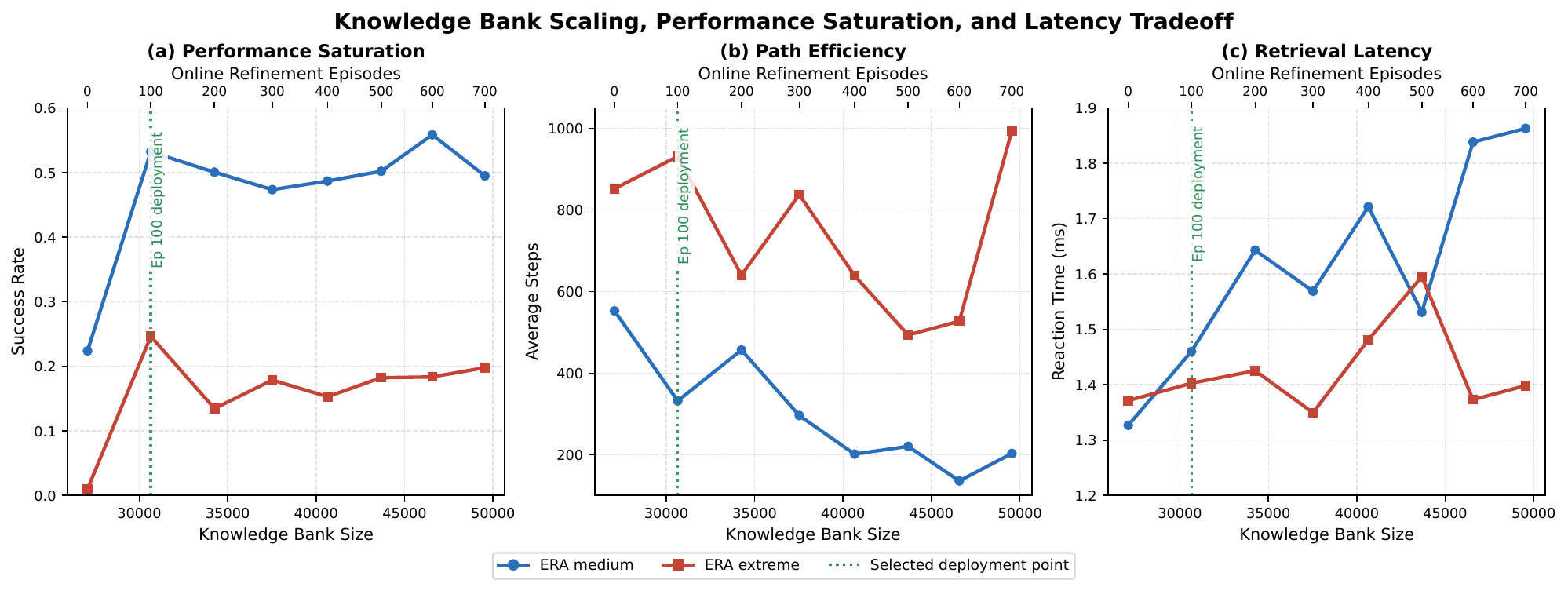}
    \caption{\textbf{Knowledge-bank scaling under the benchmark.}
    \textit{(A) Safe-progress score vs.\ KB size:} Here, the plotted ``success
    rate'' is the simulator's dense per-step safe-progress proxy rather than a
    binary goal-reached metric. Expanding the knowledge bank from the
    expert-initialized 27,075 entries (Ep.~0) to 30,650 entries (Ep.~100) yields
    the largest early gain, increasing the medium and extreme proxy scores from
    22.39\% and 1.01\% to 53.23\% and 24.64\%, respectively. Further growth to
    49,545 entries does not produce monotonic improvement, indicating saturation
    and strong sensitivity to scenario composition. \textit{(B) Effective steps
    vs.\ KB size:} Aggregate step counts remain highly variable, especially in
    extreme scenes, because timeout-heavy episodes strongly influence the
    averages; larger memory therefore improves coverage but does not by itself
    guarantee compact executions. \textit{(C) Reaction time vs.\ KB size:}
    Retrieval latency remains low across all checkpoints, ranging from 1.33 to
    1.86\,ms in medium scenes and from 1.35 to 1.60\,ms in extreme scenes.
    Because Ep.~100 achieves the strongest extreme safe-progress score while
    retaining near-peak medium performance at low latency, we use the
    $\sim$30k-entry checkpoint as the default deployed model unless otherwise
    stated.}
    \label{fig:Fig.4}
\end{figure*}

Figure~\ref{fig:Fig.4} addresses a different question from the online-learning
results in Fig.~\ref{fig:Fig.2}: not simply whether ERA changes over training,
but how performance scales with the size of the deployed knowledge bank (KB)
under the persistent-density benchmark. A key clarification is that
the success-rate curves in Fig.~\ref{fig:Fig.4} should not be read as literal
goal-completion rates. In this simulator, \texttt{success\_rate} is a dense
per-step safe-progress proxy, whereas actual episode completion is tracked by
\texttt{success\_flag} and reported separately in Table~\ref{tab:metrics}
through the goal-reach rate.

In our implementation, the bank is initialized from 27,075 expert
latent-action pairs and then enlarged online as new tuples are appended during
deployment. This produces a bank of 30,650 entries by Ep.~100 and 49,545
entries by Ep.~700. The main gain appears early: relative to the expert-only
bank, the Ep.~100 bank more than doubles the medium-difficulty proxy score
(22.39\%\,$\rightarrow$\,53.23\%) and raises the extreme-difficulty proxy score
from 1.01\% to 24.64\%. Beyond Ep.~100, however, the trend becomes clearly
non-monotonic. In medium difficulty, the proxy score fluctuates between
47.36\% and 55.85\% from Ep.~300 to Ep.~700; in extreme difficulty, none of the
later checkpoints surpasses the Ep.~100 value, instead remaining in the
13.46\%--19.79\% range.

The tables also show that bank growth affects dense action quality and
binary episode completion differently. For example, the medium goal-reach rate is
0.7 at Ep.~100 and rises to 1.0 at several later checkpoints, while the extreme
goal-reach rate varies between 0.5 and 1.0 across the same sequence. At the same
time, average final distance is very large at checkpoints with more timeout-like
episodes (e.g., 56.50 in medium and 555.71 in extreme at Ep.~100) and collapses
back to near-goal values once completion improves. These discrepancies reinforce
that the KB-size study is best interpreted as a memory-scaling analysis of
action alignment and local decision quality, not as a monotonic goal-completion
curve.

The latency tradeoff remains favorable throughout. Even the largest bank keeps
average retrieval time well below 2\,ms, so the practical cost of added memory
is modest in the current implementation. For the remainder of the paper, we
retain the Ep.~100 checkpoint unless otherwise stated, because Ep.100 provides
the highest extreme safe-progress score while maintaining low latency, subsequent
experiments use this checkpoint unless otherwise specified.

\subsubsection{Ablation Studies}
\label{ablation-studies}

\begin{figure*}
    \centering
    \includegraphics[width=\textwidth]{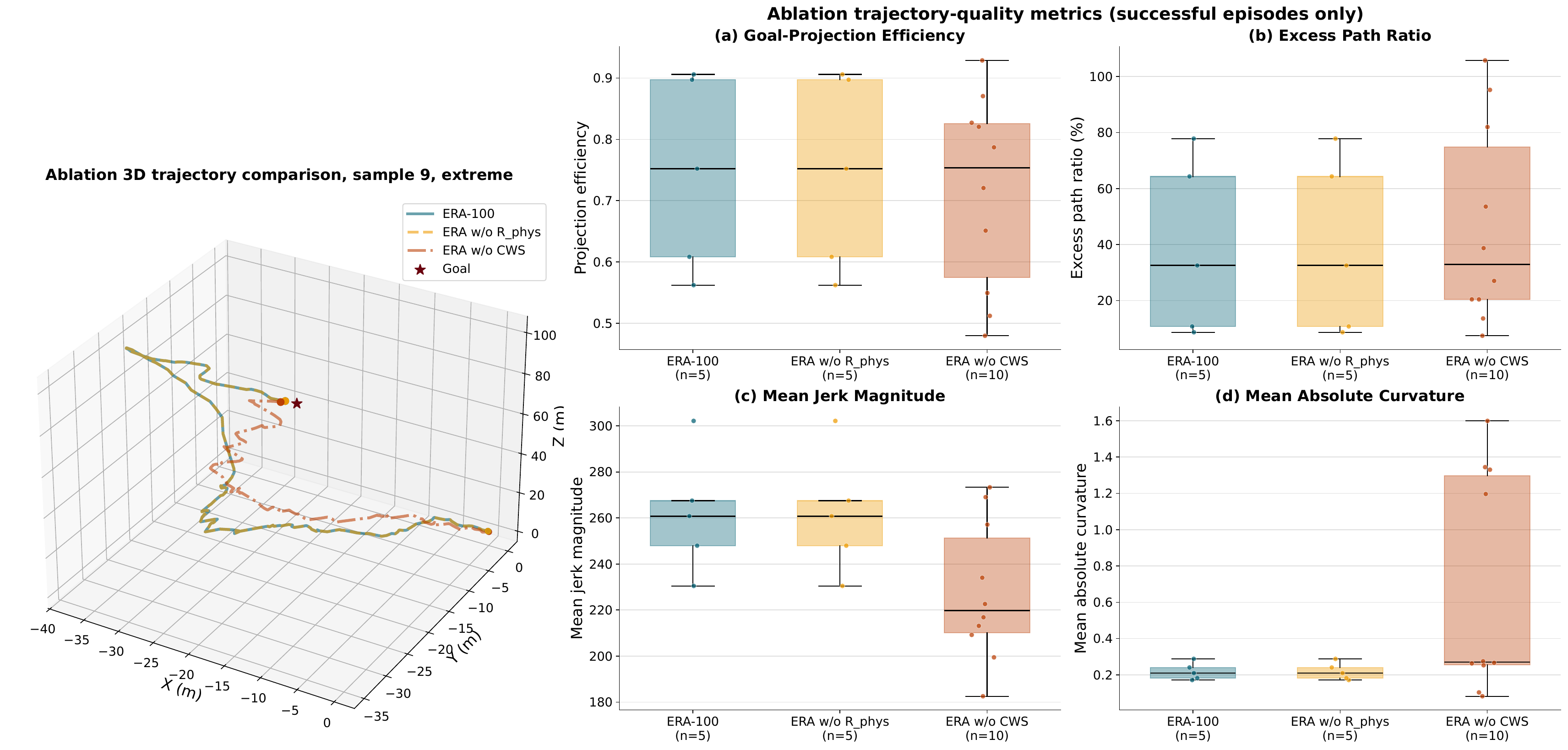}
    \caption{\textbf{Ablation analysis of latent-transition consistency and
    clustered weighted selection under extreme difficulty.}
    The left panel shows a representative benchmark episode (sample 9),
    chosen because all three variants reach the goal. In this case, ERA-100 and
    ERA w/o $R_{\mathrm{phys}}$ overlap point-for-point, so they appear as a
    single path in trajectory space, whereas removing CWS produces a visibly
    different maneuver. The right panel summarizes successful episodes only
    using goal-projection efficiency, excess path ratio, mean jerk magnitude,
    and mean absolute curvature. Together with the aggregate metrics,
    these results indicate that $R_{\mathrm{phys}}$ is at most a weak auxiliary
    regularizer in the present benchmark, while CWS has the clearer effect on
    realized trajectory structure and maneuver selection.}
    \label{fig:Fig.5}
\end{figure*}

\paragraph{Kinematic Consistency Impact}
\label{kinematic-consistency-impact}

Under the benchmark, removing $R_{\mathrm{phys}}$ produces essentially
no measurable change in deployed behavior beyond small latency fluctuations. In
both medium and extreme difficulty at Ep.~100, ERA and ERA w/o
$R_{\mathrm{phys}}$ have identical success rate, goal-reach rate, collision
rate, warning rate, average steps, and average final distance to the reported
precision; only reaction time shifts slightly (1.46 vs.\ 1.43\,ms in medium and
1.40 vs.\ 1.49\,ms in extreme). The trajectory-level ablation is even more
revealing: in the checked extreme export, ERA-100 and ERA w/o
$R_{\mathrm{phys}}$ are point-for-point identical, so the apparent disappearance
of one curve in Fig.~\ref{fig:Fig.5} reflects exact overlap rather than a
plotting artifact.

\begin{table*}
\centering
\caption{\textbf{Unified Metrics Comparison.} Upper and lower blocks denote
medium and extreme difficulty respectively under the persistent-density
benchmark. A conditional trajectory quality metrics for goal-reached episodes
only is presented in Table~\ref{tab:conditional_metrics}}
\scriptsize
\setlength{\tabcolsep}{2.5pt}
\begin{tabular}{l l c c c c c c c}
\toprule
Method & Episodes & Success Rate & Collision Rate & Warning Rate & Goal Reach Rate & Avg Steps & Reaction Time (ms) & Bank Size \\
\midrule
\multirow{8}{*}{\textbf{ERA}} 
& 0   & 0.2239 & 0.0157 & 0.7604 & 1.0 & 552.6 & 1.3268 & 27075 \\
& 100 & 0.5323 & 0.0502 & 0.4175 & 0.7 & 331.9 & 1.4603 & 30650 \\
& 200 & 0.5006 & 0.0603 & 0.4390 & 0.6 & 456.6 & 1.6430 & 34252 \\
& 300 & 0.4736 & 0.0789 & 0.4475 & 0.9 & 296.0 & 1.5690 & 37521 \\
& 400 & 0.4869 & 0.0825 & 0.4306 & 1.0 & 201.4 & 1.7215 & 40648 \\
& 500 & 0.5021 & 0.0826 & 0.4153 & 1.0 & 220.4 & 1.5313 & 43677 \\
& 600 & 0.5585 & 0.0713 & 0.3702 & 1.0 & 135.7 & 1.8384 & 46570 \\
& 700 & 0.4949 & 0.0907 & 0.4144 & 1.0 & 202.9 & 1.8631 & 49545 \\
\textbf{ERA w/o $R_{\mathbf{phys}}$} 
& 100 & 0.5323 & 0.0502 & 0.4175 & 0.7 & 331.9 & 1.4343 & 30650 \\
\textbf{ERA w/o CWS}
& 100 & 0.4010 & 0.0500 & 0.5490 & 1.0 & 175.2 & 1.5569 & 28034 \\
\addlinespace
\textbf{VPF Expert}
& - & 0.6021 & 0.0000 & 0.3979 & 1.0 & 443.9 & 0.5718 & N/A \\
\textbf{BC-IL}
& - & 0.5256 & 0.0422 & 0.4322 & 0.3 & 1159.2 & 0.5794 & N/A \\
\textbf{Acados}
& - & 0.3273 & 0.0350 & 0.6376 & 1.0 & 248.7 & 29.0837 & N/A \\
\textbf{PPO}
& - & 0.2003 & 0.0082 & 0.7915 & 1.0 & 742.3 & 2.7772 & N/A \\
\addlinespace
\textbf{ERA 25\%}
& 100 & 0.4144 & 0.0676 & 0.5179 & 1.0 & 192.6 & 1.6180 & 10064 \\
\textbf{BC-IL 25\%}
& - & 0.4896 & 0.0577 & 0.4527 & 1.0 & 154.4 & 0.5712 & N/A \\
\textbf{ERA 50\%}
& 100 & 0.3658 & 0.0515 & 0.5827 & 1.0 & 197.5 & 1.2994 & 15812 \\
\textbf{BC-IL 50\%}
& - & 0.3991 & 0.0731 & 0.5278 & 1.0 & 224.5 & 0.5809 & N/A \\
\midrule
\multirow{8}{*}{\textbf{ERA}} 
& 0   & 0.0101 & 0.0225 & 0.9674 & 1.0 & 852.2 & 1.3714 & 27075 \\
& 100 & 0.2464 & 0.1342 & 0.6195 & 0.5 & 931.4 & 1.4028 & 30650 \\
& 200 & 0.1346 & 0.1783 & 0.6871 & 1.0 & 640.3 & 1.4254 & 34252 \\
& 300 & 0.1787 & 0.1716 & 0.6497 & 0.9 & 837.6 & 1.3495 & 37521 \\
& 400 & 0.1530 & 0.1907 & 0.6563 & 0.8 & 638.9 & 1.4811 & 40648 \\
& 500 & 0.1824 & 0.1550 & 0.6626 & 1.0 & 493.5 & 1.5953 & 43677 \\
& 600 & 0.1836 & 0.1737 & 0.6428 & 1.0 & 527.3 & 1.3731 & 46570 \\
& 700 & 0.1979 & 0.1659 & 0.6361 & 0.9 & 994.6 & 1.3985 & 49545 \\
\textbf{ERA w/o $R_{\mathbf{phys}}$}
& 100 & 0.2464 & 0.1342 & 0.6195 & 0.5 & 931.4 & 1.4945 & 30650 \\
\textbf{ERA w/o CWS}
& 100 & 0.0976 & 0.1225 & 0.7799 & 1.0 & 277.1 & 1.4808 & 28034 \\
\addlinespace
\textbf{VPF Expert}
& - & 0.1679 & 0.0188 & 0.8134 & 1.0 & 940.9 & 0.6212 & N/A \\
\textbf{BC-IL}
& - & 0.1347 & 0.1472 & 0.7181 & 0.5 & 1345.2 & 0.5552 & N/A \\
\textbf{Acados}
& - & 0.0320 & 0.0925 & 0.8755 & 1.0 & 306.6 & 26.0540 & N/A \\
\textbf{PPO}
& - & 0.0075 & 0.0233 & 0.9693 & 1.0 & 894.0 & 2.2571 & N/A \\
\addlinespace
\textbf{ERA 25\%}
& 100 & 0.0984 & 0.1552 & 0.7465 & 0.9 & 663.5 & 1.3598 & 10064 \\
\textbf{BC-IL 25\%}
& - & 0.1336 & 0.1625 & 0.7039 & 1.0 & 194.1 & 0.6181 & N/A \\
\textbf{ERA 50\%}
& 100 & 0.0928 & 0.1051 & 0.8021 & 0.3 & 1459.4 & 1.3606 & 15812 \\
\textbf{BC-IL 50\%}
& - & 0.0987 & 0.1449 & 0.7564 & 1.0 & 217.5 & 0.5478 & N/A \\
\bottomrule
\end{tabular}
\label{tab:metrics}
\end{table*}

\paragraph{Weighted Clustering Analysis}
\label{weighted-clustering-analysis}

CWS has a substantially larger behavioral footprint than
$R_{\mathrm{phys}}$. Removing CWS changes the realized trajectory geometry
visibly in the representative extreme episode and degrades the dense
safe-progress metric in both difficulty settings (Table~\ref{tab:metrics}):
in medium difficulty, the score drops from 0.5323 to 0.4010 and the warning
rate rises from 0.4175 to 0.5490; in extreme difficulty, the score drops from
0.2464 to 0.0976 and the warning rate rises from 0.6195 to 0.7799. These shifts
indicate that CWS plays an important role in maintaining decisional
consistency and safety margin during multimodal conflict resolution.

At the same time, the ablation also shows that binary completion and trajectory
quality can separate. According to Table~\ref{tab:metrics}, the variant without
CWS still attains a goal-reach rate of 1.0 in both medium and extreme
difficulty, despite producing visibly different paths and weaker stepwise
safe-progress statistics. Moreover, Fig.~\ref{fig:Fig.5} proved that without
the CWS, ERA perform lower in all the four metrics in the bosplots. We
therefore interpret CWS primarily as a maneuver selection mechanism that shapes
how the controller resolves ambiguous local retrievals, rather than as a simple
on/off success booster. This reading is consistent with the successful-episode
boxplots and the 3D trajectory views: among the tested components, removing CWS
produces the clearest change in path structure, whereas removing
$R_{\mathrm{phys}}$ leaves the deployed trajectories nearly unchanged.

\subsection{Baseline Comparison and Deployment Efficiency}
\label{baseline-comparison-and-deployment-efficiency}

\subsubsection{Comparative Evaluation}
\label{comparative-evaluation}

Table~\ref{tab:metrics} compares ERA against four representative navigation
paradigms under a shared fixed-difficulty evaluation protocol in Isaac Sim:
the VPF expert teacher, a behavioral cloning baseline (BC-IL), a PPO-based
policy-gradient baseline \cite{schulman2017proximal}, and an Acados-based
model predictive controller \cite{verschueren2022acados}. These baselines were
chosen to span four distinct design philosophies for autonomous navigation:
explicit expert control, parametric imitation learning, reinforcement learning,
and online optimization.

All methods are evaluated through the same rollout interface using identical goal
sampling, persistent-density intruder management, and the same episode-level
metrics: dense safe-progress, collision rate, warning rate, goal-reach rate (goal
reached within 2000 steps), average steps, average final distance, and reaction
time. All controllers are deployed through a unified interface in Isaac Sim with
online adaptation disabled.

Under this protocol, every policy within a benchmark block encounters the same
seeded goal distributions and identical adversarial intruder behaviors,
including collision-course, near-miss, crossing, and multi-conflict encounter
patterns generated from a shared difficulty profile. The intent of this design
is not merely to compare task completion, but to test how different navigation
paradigms trade off local safety progress, actual goal completion, trajectory
compactness, and deployment-time responsiveness under matched multi-agent
stress.

\begin{figure}
    \centering
    \includegraphics[width=\columnwidth]{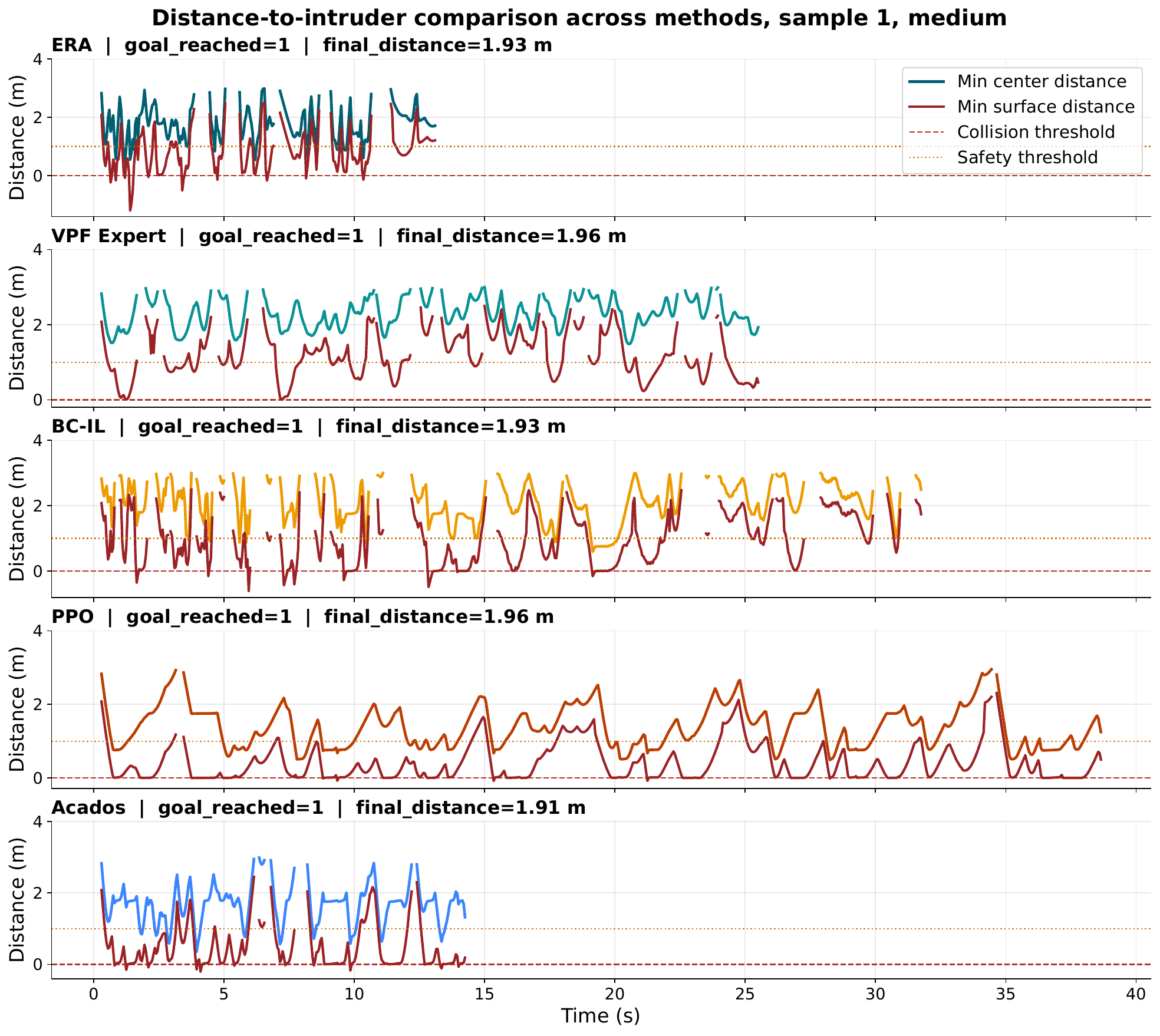}
    \caption{\textbf{Distance-to-intruder evolution for five navigation methods
    on the same evaluation episode (sample 1, medium difficulty).}
    Each subplot reports one method and shows the minimum center-to-center
    distance between the ego drone and the detected intruders, together with
    the minimum surface clearance after accounting for body radius. The dashed
    horizontal line marks the collision boundary at 0\,m, while the dotted
    horizontal line marks the 1\,m safety threshold. Larger positive surface
    distances indicate safer motion, values between 0 and 1\,m indicate
    near-conflict operation, and negative values indicate geometric overlap.
    In this sample, all five methods reach the goal, so the figure isolates
    differences in transient safety margin and conflict handling rather than
    simple success versus failure.}
    \label{fig:Fig.6}
\end{figure}

\begin{figure}
    \centering
    \includegraphics[width=\columnwidth]{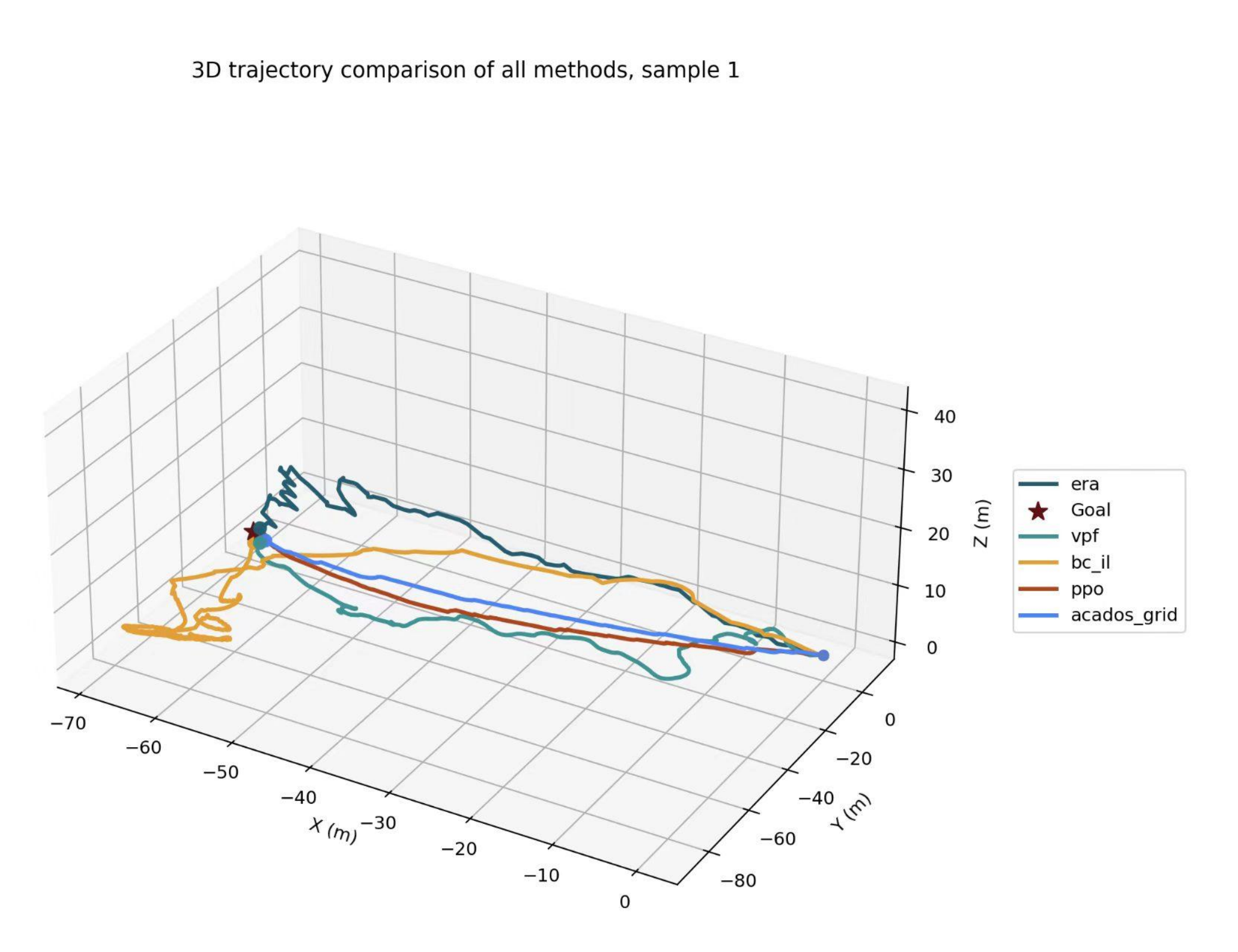}
    \caption{\textbf{Three-dimensional trajectory comparison for five navigation
    methods on the same evaluation episode (sample 1, medium difficulty).}
    Each curve shows the full spatial path executed by one controller under the
    same seeded goal and adversarial intruder scenario, allowing direct
    geometric comparison of avoidance behavior under matched conditions. The
    shared start location, terminal position of each method, and common goal
    location are visualized in the same frame. Because all methods in this
    sample reach the goal, differences in path curvature, detour magnitude, and
    final approach reflect distinct maneuver-selection strategies rather than
    binary task completion alone.}
    \label{fig:Fig.7}
\end{figure}

In medium difficulty, ERA-100 attains the highest dense safe-progress score among
the learned and optimization-based baselines at 0.5323, narrowly above BC-IL
(0.5256), while remaining far faster than Acados (1.46\,ms vs.\ 29.08\,ms) and
substantially stronger than PPO (0.2003). In extreme difficulty, ERA-100 achieves
the strongest dense safe-progress score of all methods at 0.2464, exceeding VPF
(0.1679), BC-IL (0.1347), Acados (0.0320), and PPO (0.0075). Dense safe-progress
and binary goal completion measure different aspects of performance. For example,
in medium difficulty VPF reaches the goal in all episodes whereas ERA succeeds in
7/10. In extreme difficulty, ERA and BC-IL each achieve 5/10 goal completion
despite different dense safe-progress scores. Table~\ref{tab:metrics} therefore
reports both metrics to distinguish continuous navigation quality from final task
completion.

\paragraph{VPF Teacher}
\label{vpf-teacher}

The VPF teacher serves as the expert reference policy from which demonstrations
are collected. It achieves a goal-reach rate of 1.0 in both medium and extreme
difficulty and also delivers the best collision rates among the nontrivial
baselines. ERA does not surpass VPF on these binary reliability measures but
appears in dense action quality under hard scenarios: in extreme difficulty,
ERA-100 achieves a much higher safe-progress score than VPF (0.2464 vs.\ 0.1679),
indicating more favorable stepwise behavior despite lower episode completion.

\begin{figure*}
    \centering
    \includegraphics[width=1.3\columnwidth]{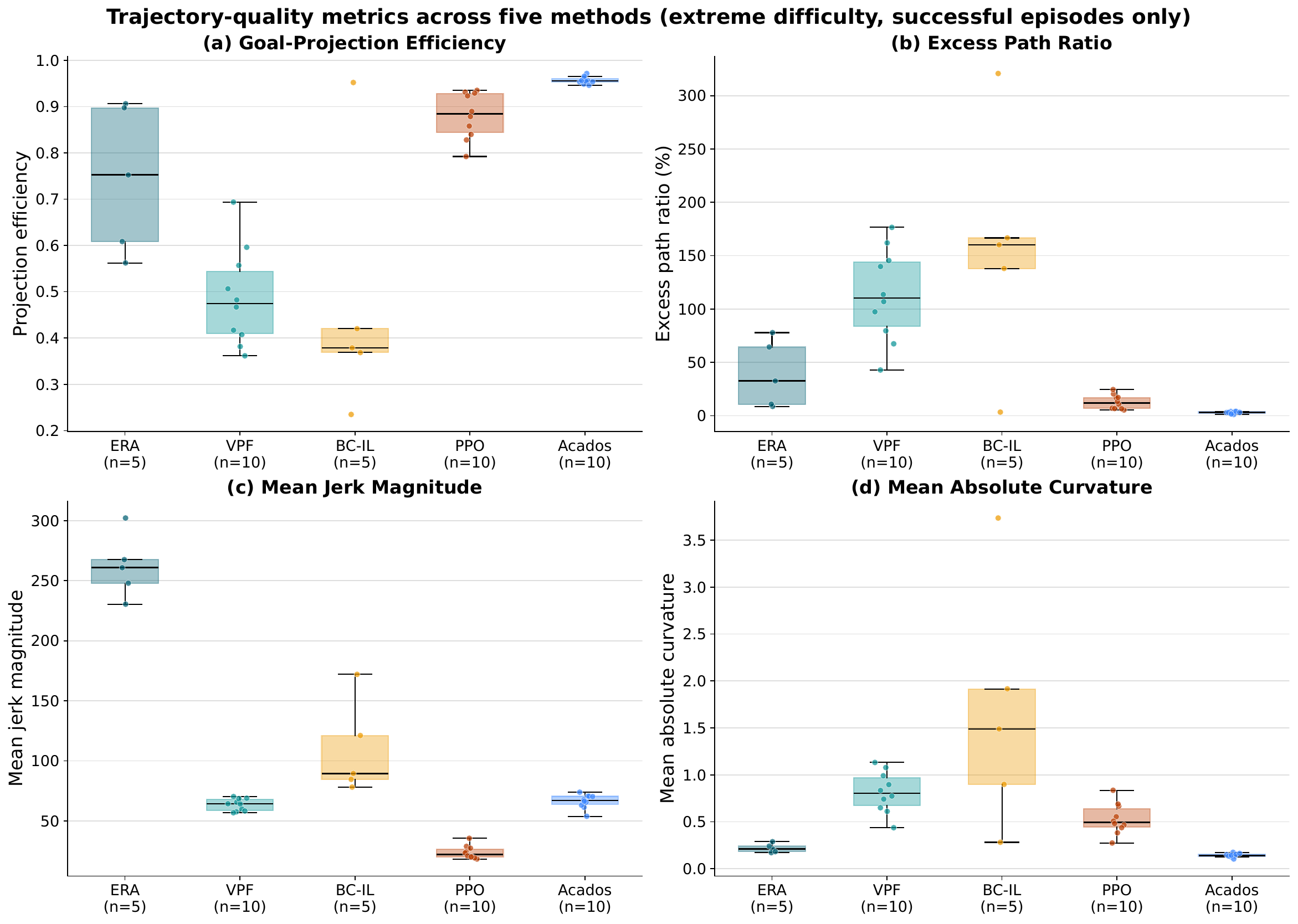}
    \caption{\textbf{Distribution of trajectory-quality metrics across five
    methods under extreme difficulty.}
    The boxplots summarize episode-level performance over successful runs only.
    \textit{(A) Goal-projection efficiency:} higher values indicate that a
    larger fraction of the traveled motion contributes directly toward the goal.
    \textit{(B) Excess path ratio:} lower values indicate less path overhead
    relative to the straight-line start-to-goal distance. \textit{(C) Mean jerk
    magnitude:} lower values indicate smoother control commands and reduced
    aggressive motion changes. \textit{(D) Mean absolute curvature:} lower
    values indicate smoother, less sharply turning trajectories. Because only
    goal-reaching episodes are included, this figure isolates trajectory quality
    conditional on successful completion rather than overall benchmark
    robustness.}
    \label{fig:Fig.8}
\end{figure*}

\paragraph{Behavioral Cloning (BC/IL)}
\label{behavioral-cloning}

BC-IL provides the closest learning baseline because both methods are trained
from the same expert demonstrations. In the full-data benchmark, BC-IL remains
competitive in medium difficulty but trails ERA in extreme dense safe-progress
(0.1347 vs. 0.2464). Under the 25\% and 50\% data settings, BC-IL slightly
outperforms ERA in several cases, indicating that ERA's advantage becomes clearer
once the knowledge bank contains sufficient experience.

\paragraph{Proximal Policy Optimization (PPO)}
\label{proximal-policy-optimization}

PPO uses a fixed-width observation representation derived from the event list and
is trained by policy optimization. It achieves the lowest dense safe-progress
scores under both difficulty settings while maintaining relatively high goal
completion. As shown in Fig~\ref{fig:Fig.8}, PPO produces efficient successful
trajectories but accumulates substantially higher warning rates and longer
rollouts.

\paragraph{Model Acados (MPC)}
\label{model-acados}

Acados provides a model-based optimization baseline. It achieves perfect goal
completion but requires approximately 26–29 ms per decision, compared with
approximately 1.5 ms for ERA. Dense safe-progress is also consistently lower than
ERA in both benchmark settings.

\paragraph{Conditional Trajectory Quality}
\label{conditional-trajectory-quality}

\begin{table}
\centering
\caption{\textbf{Conditional Trajectory Quality Metrics (Goal-Reached Episodes
Only).}
Metrics are computed only over episodes that successfully reached the goal,
isolating trajectory efficiency and safety quality after task completion.
Upper and lower blocks denote medium and extreme difficulty respectively.}
\scriptsize
\setlength{\tabcolsep}{2.5pt}
\begin{tabular}{l l c c c c }
\toprule
Method & Episodes & Success Rate & Collision Rate & Warning Rate & Avg Steps \\
\midrule
\multirow{8}{*}{\textbf{ERA}}
& 0   & 0.2239 & 0.0157 & 0.7604 & 552.6 \\
& 100 & 0.4549 & 0.0587 & 0.4864 & 157.7 \\
& 200 & 0.3977 & 0.0733 & 0.5290 & 144.0 \\
& 300 & 0.4572 & 0.0831 & 0.4597 & 208.8 \\
& 400 & 0.4869 & 0.0825 & 0.4306 & 201.4 \\
& 500 & 0.5021 & 0.0826 & 0.4153 & 220.4 \\
& 600 & 0.5585 & 0.0713 & 0.3702 & 135.7 \\
& 700 & 0.4949 & 0.0907 & 0.4144 & 202.9 \\
\addlinespace
\textbf{VPF}
& - & 0.6021 & 0.0000 & 0.3979 & 443.9 \\
\textbf{BC-IL}
& - & 0.3769 & 0.0585 & 0.5645 & 256.7 \\
\textbf{Acados}
& - & 0.3273 & 0.0350 & 0.6376 & 248.7 \\
\textbf{PPO}
& - & 0.2003 & 0.0082 & 0.7915 & 742.3 \\
\midrule
\multirow{8}{*}{\textbf{ERA}}
& 0   & 0.0101 & 0.0225 & 0.9674 & 852.2 \\
& 100 & 0.0658 & 0.1653 & 0.7689 & 215.8 \\
& 200 & 0.1346 & 0.1783 & 0.6871 & 640.3 \\
& 300 & 0.1785 & 0.1735 & 0.6480 & 711.9 \\
& 400 & 0.1500 & 0.1949 & 0.6551 & 305.5 \\
& 500 & 0.1824 & 0.1550 & 0.6626 & 493.5 \\
& 600 & 0.1836 & 0.1737 & 0.6428 & 527.3 \\
& 700 & 0.1938 & 0.1689 & 0.6373 & 887.8 \\
\addlinespace
\textbf{VPF}
& - & 0.1679 & 0.0188 & 0.8134 & 940.9 \\
\textbf{BC-IL}
& - & 0.0919 & 0.0987 & 0.8094 & 741.6 \\
\textbf{Acados}
& - & 0.0320 & 0.0925 & 0.8755 & 306.6 \\
\textbf{PPO}
& - & 0.0075 & 0.0233 & 0.9693 & 894.0 \\
\bottomrule
\end{tabular}
\label{tab:conditional_metrics}
\end{table}

Table~\ref{tab:conditional_metrics} provides an additional diagnostic computed 
only over goal-reaching episodes. Unlike the primary benchmark in
Table~\ref{tab:metrics}, which evaluates overall task reliability including 
completion frequency, this analysis isolates how efficiently and safely each
controller behaves once successful completion occurs. The objective is to
separate execution quality from failure frequency and determine whether weaker 
aggregate metrics are driven by poor trajectory execution or simply by lower 
completion rates.

Overall, ERA achieves the strongest dense safe-progress under extreme difficulty
while maintaining low deployment latency. By contrast, VPF provides the highest
binary robustness, and BC-IL, PPO, and Acados each exhibit complementary strengths
in imitation, optimization, or trajectory quality.

\subsubsection{Inference Latency}
\label{inference-latency}

To assess whether ERA is practically deployable \cite{lee2025performance}, we
evaluate the deployed controller on an NVIDIA Jetson Orin Nano Super (4GB). The
deployment audit loads a frozen checkpoint together with a frozen memory snapshot
and measures both runtime latency and memory usage under GPU execution.

\begin{table}
\centering
\caption{Latency Breakdown on NVIDIA Jetson Orin Nano}
\label{tab:latency}
\scriptsize
\setlength{\tabcolsep}{4pt}
\renewcommand{\arraystretch}{1.05}
\resizebox{\columnwidth}{!}{
\begin{tabular}{lcc}
\toprule
\textbf{Component} & \textbf{FP32 Latency (ms)} & \textbf{FP16 Latency (ms)} \\
\midrule
Encoder & 0.898 & 0.970 \\
Knowledge Bank & 2.546 & 3.418 \\
Stabilizer & 0.556 & 0.767 \\
Action Fusion & 1.183 & 1.365 \\
\midrule
\textbf{End-to-End Total} & \textbf{7.649} & \textbf{9.414} \\
\bottomrule
\end{tabular}
}
\end{table}

Latency is evaluated using the deployment benchmark, where each component is
executed after a 30-iteration warm-up phase, and the mean runtime is reported
over 200 synchronized GPU iterations. This protocol removes one-time
initialization overhead and ensures that the reported measurements reflect
steady-state inference performance rather than transient launch variability.
The benchmark independently profiles the encoder forward pass, knowledge bank
retrieval, Lyapunov stability filtering, final clustered action fusion, and the
complete end-to-end inference pipeline.

As shown in Table~\ref{tab:latency}, ERA achieves an end-to-end latency of
$7.649$ ms in FP32 mode, corresponding to an effective control frequency of
approximately $130.7$ Hz \cite{mellinger2011minimum}. This demonstrates that ERA's
retrieval, stability, and fusion modules can operate directly within a real-time
flight loop rather than being limited to offline planning. The component-level
breakdown is also informative: knowledge bank retrieval represents the largest
computational cost, while the Lyapunov-based stability filter adds only $0.556$
ms of overhead, indicating that safety verification is not the primary bottleneck.

\subsubsection{Resource Efficiency and Memory Footprint}
\label{resource-efficiency-and-memory-footprint}

Memory usage is evaluated through both static deployment footprint and runtime
GPU allocation. The deployed checkpoint remains compact, requiring only $15.77$
MB in total, consisting of 0.33 MB of model parameters and a 15.43 MB knowledge
bank, leaving substantial memory headroom on the evaluated 4 GB Jetson platform.
Runtime profiling further shows that knowledge bank storage scales approximately
linearly with the number of stored experiences, providing predictable memory
growth for larger repositories. Moreover, the Lyapunov stability filter introduces
negligible deployment overhead, adding only 0.556 ms in FP32 mode without
materially affecting memory consumption. These results suggest that ERA's
computational requirements remain lightweight, with practical deployment
constraints more likely to arise from the surrounding perception and
state-estimation pipeline than from the retrieval controller itself.

\subsubsection{Knowledge Bank Scalability}
\label{knowledge-bank-scalability}

\begin{table}
\centering
\caption{\textbf{Scalability of the ERA Knowledge Bank under retrieval-time
benchmarking.}
Each checkpoint stores a larger non-parametric knowledge bank, and we report how
this growth affects in-memory payload size and retrieval latency. `Bank Size` is
the number of stored memory entries, `Payload Mem.` is the tensor footprint
of the stored arrays, `Mean Lat.` is the average time per retrieval call, and
`p95 Lat.` is the 95th-percentile retrieval time. Measurements were obtained on
a cuda device using the deployed ERA retrieval kernel with top-$k=5$,
evaluated over $N=256$ sampled query latents per checkpoint. As the bank grows
from 27,075 to 49,545 entries, memory increases predictably while retrieval
latency remains within real-time control bounds.}
\scriptsize
\setlength{\tabcolsep}{5pt}
\begin{tabular}{l c c c c}
\toprule
Episodes & Bank Size & Payload Mem. (MB) & Mean Lat. (ms) & p95 Lat. (ms) \\
\midrule
Pretrained & 27,075 & 13.63 & 4.126 & 4.819 \\
100 & 30,650 & 15.43 & 4.982 & 5.052 \\
200 & 34,252 & 17.25 & 5.490 & 5.547 \\
300 & 37,521 & 18.89 & 5.803 & 5.854 \\
400 & 40,648 & 20.47 & 6.199 & 6.244 \\
500 & 43,677 & 21.99 & 6.604 & 6.662 \\
600 & 46,570 & 23.45 & 6.989 & 7.055 \\
700 & 49,545 & 24.95 & 7.387 & 7.451 \\
\bottomrule
\end{tabular}
\label{tab:inference_benchmarks}
\end{table}

\begin{figure}
    \centering
    \includegraphics[width=\columnwidth]{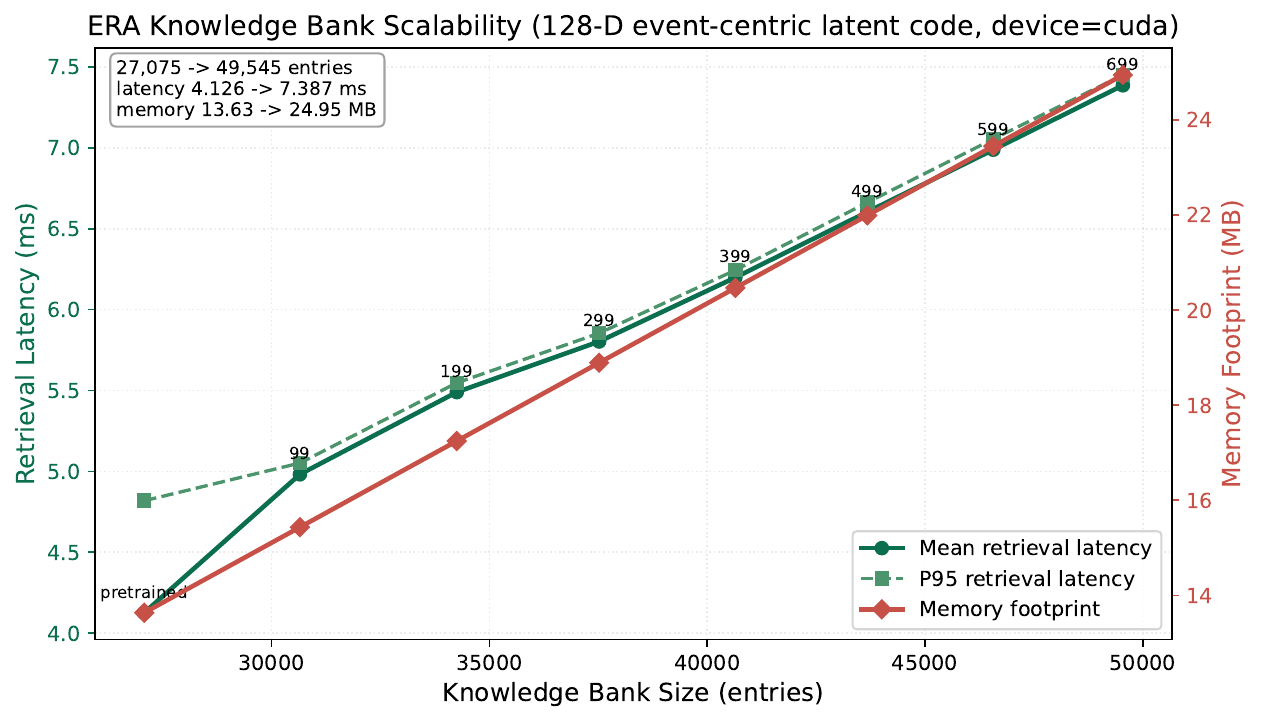}
    \caption{\textbf{Dual-axis scalability profile of the ERA knowledge bank.}
    The horizontal axis shows knowledge bank size across training checkpoints,
    the left vertical axis reports retrieval latency in milliseconds, and the
    right vertical axis reports payload memory in megabytes. Latency is measured
    with the deployed ERA retrieval kernel using top-$k=5$ nearest-neighbor
    search over $N=256$ sampled query latents on a \texttt{cuda} device, while
    payload memory is computed from the stored \texttt{latents},
    \texttt{actions}, and \texttt{reliability} tensors. As the bank grows from
    27,075 to 49,545 entries, memory increases nearly linearly, whereas
    retrieval latency rises more gradually and remains within real-time
    operating bounds.}
    \label{fig:Fig.9}
\end{figure}

Table~\ref{tab:inference_benchmarks} and Fig.~\ref{fig:Fig.9} evaluate how ERA's
deployed knowledge bank scales as additional experience is accumulated during
training. The analysis characterizes both storage overhead and runtime retrieval
cost using four metrics: Bank Size, Payload Memory (MB), Mean Latency (ms), and
p95 Latency (ms). Benchmarks are performed directly on saved checkpoint
snapshots. For each checkpoint, the system loads the stored latents, actions, and
reliability tensors, computes the effective payload memory, and evaluates a
deployment-aligned per-query retrieval benchmark based on nearest-neighbor
search, top-$k=5$ retrieval, and weighted action aggregation over $N=256$
sampled query latents.

\section{Discussion}
\label{discussion}

The experimental results indicate that the central value of ERA is to demonstrate
the viability of the proposed perspective about embodied decision-making. The
experiments demonstrate that the proposed framework explicitly exposes the prior
experience supporting each retrieved action. Moreover, ERA jointly emphasizes
interpretability, real-time responsiveness, safety-aware action selection, and
deployability on resource-constrained edge hardware. Compared to fully parametric
policies and online optimization methods, ERA stores decision evidence explicitly
in the knowledge bank. This allows each action to be traced back to similar
historical events and their associated executions, shifting part of the real-time
decision burden from iterative solving to efficient retrieval and selection over
stored event-action experience, making it particularly suited for regimes like
high-frequency control loops on edge platforms.

One important insight for the observed collision rate is the finite coverage of
the current knowledge bank. Because ERA derives actions through explicit
retrieval rather than implicit parameterized policies, scenarios lacking
sufficiently similar prior experiences are expected to present greater challenges
than well-represented situations compared to BC-IL and VPF Expert. This reflects
an inherent trade-off of retrieval-based decision making. Unarguable, ERA
behaves competitively for situations that are sufficiently represented by prior
experience, while failures concentrate in challenging or insufficiently covered
scenarios. Those uncovered scenarios are identifiable, provided with a trasparent
basis for future confidence-aware intervention or human-in-the-loop operation. In
other words, those remaining collisions primarily arise in scenarios
insufficiently represented by the current knowledge bank, rather than from
instability of the retrieval mechanism itself. ERA avoids the significant
exploration costs of PPO and the online solving burden of Acados by reusing
historical event-action experience. This trade-off aligns with the constraints of
the current benchmark, which prioritizes rapid response, trajectory compactness,
and stable behavior under dynamic interaction. This deployment claim is further
supported by the scalability benchmark, which shows that as the knowledge bank
grows from 27,075 to 49,545 entries, payload memory increases predictably from
13.63 MB to 24.95 MB, while per-query retrieval latency rises more gradually from
4.126 ms to 7.387 ms. This behavior is consistent with a saturation effect, in
which additional demonstrations increasingly densify existing regions of the
knowledge bank rather than introducing genuinely new maneuver mode.

The ablation results further clarify the role of the proposed components. The
benchmark suggests that $R_{\mathrm{phys}}$ primarily contributes to representation
alignment rather than directly affecting task-level performance, suggesting
that it acts at most as a latent-transition consistency regularizer. During
training, it encourages the predicted latent dynamics to remain compatible with
observed transitions, improving representation coherence and interpretability.
Once the retrieval controller and expert memory are fixed, however, this effect
appears secondary to deployment performance. By contrast, removing CWS changes the
realized trajectory geometry more visibly and lowers the dense safe-progress
metric in both medium and extreme difficulty, even though goal-reach frequency is
not a monotonic function of the full method in the benchmark. Therefore, CWS is
best understood as a maneuver-selection mechanism that improves decisional
consistency in multimodal conflicts, whereas $R_{\mathrm{phys}}$ appears secondary
in the present evaluation, which is heuristically designed.

Several limitations remain. First, the current retrieval implementation scales
approximately linearly with knowledge bank size in both storage and query cost.
Although the measured growth remains moderate over the tested range, with
retrieval latency increasing from 4.126 ms to 7.387 ms as the bank expands from
27,075 to 49,545 entries, larger repositories will eventually require indexed
retrieval, bank compression, or active sample selection. Second, the
current benchmark, while useful for observing multimodal conflicts, does not yet
isolate a "canonical dilemma" (e.g., a pure fork benchmark) to prove the CWS
mechanism in a standalone way. Third, the current validation still remains
largely based on simulation and simplified UAV geometry; more realistic
perception noise, complex UAV dynamics, and real-flight validation are necessary
to assess robustness under real-world distribution shifts.

\section{Conclusion and Future Works}
\label{conclusion-and-future-works}

This paper presented ERA, an event-centric retrieval-action framework for
interpretable embodied decision-making that bridge the gap between black-box
control and traditional optimization. By abstracting dynamic environments into
structured event lists and retrieving prior experience from a knowledge bank,
ERA generates actions through reliability weighting, clustered selection (CWS),
and Lyapunov-inspired stability filtering. Experiments in Isaac Sim UAV
navigation demonstrate that ERA provides a practical trade-off among dense
safe-progress behavior, trajectory structure, and latency. Specifically, on a
Jetson Orin Nano, ERA achieves an update frequency of 130.7 Hz while
maintaining a memory footprint compatible with edge-hardware constraints.
Moreover, the dedicated scalability analysis shows that this favorable systems
profile persists as the knowledge bank expands: across checkpoints, payload
memory grows predictably while retrieval latency remains within real-time
operating bounds. These results suggest that explicit retrieval provides a viable
alternative to fully parameterized embodied policies for safety-critical embodied
decision making. While the current implementation remains limited by knowledge
bank coverage, the experiments demonstrate that interpretable memory-based
reasoning can support real-time deployment without requiring all prior experience
to be compressed into neural network parameters. Rather than replacing existing
retrieval, structured memory, and transparent reasoning jointly from the basis
of embodied decision making.

Future work will focus on several key directions to enhance the scalability and
robustness of the framework:
\begin{itemize}
\item \textbf{Efficiency:} Implementing indexed retrieval
\cite{jegou2010product, johnson2019billion}, knowledge bank compression
\cite{katraouras2026memorybankcompressioncontinual}, and sample replacement
strategies \cite{chrysakis2020online, chaudhry2019tiny} to preserve the
favorable memory-latency scaling observed here as experience grows beyond the
current benchmark range.
\item \textbf{Context:} Incorporating richer contextual event attributes to
improve the nuance of the latent representations.
\item \textbf{Validation:} Evaluating ERA under more realistic perception noise,
complex UAV dynamics, and real-world flight conditions to move beyond geometric
obstacle avoidance toward general physically grounded decision-making.
\end{itemize}

\bibliographystyle{IEEEtran}
\bibliography{refs}

\end{document}